# Predicting Liquidity Coverage Ratio with Gated Recurrent Units: A Deep Learning Model for Risk Management

Zhen Xu
Independent Researcher
Shanghai, China

Jingming Pan
University of Pennsylvania
Philadelphia, USA

Siyuan Han
Columbia University
New York, USA

Hongju Ouyang
Columbia University
New York City, USA

Yuan Chen
Rice University
Houston, USA

Mohan Jiang*
New York University
New York, USA

*Abstract*—With the global economic integration and the high interconnection of financial markets, financial institutions are facing unprecedented challenges, especially liquidity risk. This paper proposes a liquidity coverage ratio (LCR) prediction model based on the gated recurrent unit (GRU) network to help financial institutions manage their liquidity risk more effectively. By utilizing the GRU network in deep learning technology, the model can automatically learn complex patterns from historical data and accurately predict LCR for a period of time in the future. The experimental results show that compared with traditional methods, the GRU model proposed in this study shows significant advantages in mean absolute error (MAE), proving its higher accuracy and robustness. This not only provides financial institutions with a more reliable liquidity risk management tool but also provides support for regulators to formulate more scientific and reasonable policies, which helps to improve the stability of the entire financial system.

I. INTRODUCTION

In today's context of global economic integration and highly interconnected financial markets, financial institutions are facing unprecedented challenges, especially liquidity risk [1]. Liquidity risk refers to the risk that banks or other financial institutions cannot quickly obtain sufficient funds at a reasonable cost to meet their short-term debt or operational needs [2]. Once this risk gets out of control, it may not only cause individual institutions to fall into financial difficulties but may also trigger a chain reaction, causing serious impacts on the entire financial system and even the macroeconomy [3].

In order to better manage and monitor liquidity risk, international regulators such as the Basel Committee have proposed a key indicator, the Liquidity Coverage Ratio (LCR) [4]. LCR aims to ensure that banks have sufficient high-quality liquid assets (HQLA) to cover the expected net cash outflow within 30 days under a stress scenario. By maintaining a high LCR level, banks can be better prepared to deal with sudden funding needs, thereby enhancing their ability to resist short-term liquidity crises. However, accurately predicting LCR is not easy [5]. Traditional statistical methods usually perform linear regression or time series analysis based on historical data, but these methods often fail to provide accurate prediction results when faced with complex market environments, emergencies, and nonlinear economic relationships [6].

With the development of big data technology and the advancement of artificial intelligence algorithms, deep learning models have gradually become an effective tool for solving complex prediction problems. In particular, recurrent neural networks (RNNs) have shown unique advantages in processing time series data. As a special type of RNN structure, the gated recurrent unit (GRU) can effectively capture long-term dependencies in time series by introducing a gating mechanism to control information flow and has a simpler structure and higher training efficiency than the long short-term memory (LSTM) network [7]. These characteristics of GRU make it an ideal choice for improving the LCR prediction model [8].

The significance of using the GRU network to improve the LCR prediction model is to improve the prediction accuracy. By capturing the complex patterns and dynamic changes in the time series, the GRU network can provide more accurate LCR prediction results than traditional methods, helping financial institutions identify potential liquidity risks earlier. Accurate LCR prediction helps financial institutions formulate corresponding risk management strategies in advance, such as adjusting asset-liability structure, increasing high-quality liquid asset reserves, etc., so as to effectively reduce liquidity risk. For regulators, the LCR prediction model based on the GRU network can provide them with a more reliable liquidity risk assessment tool, help formulate more scientific and reasonable regulatory policies, and further enhance the stability of the financial system.

In addition, the LCR prediction model based on the GRU network can also help financial institutions optimize resource allocation and improve capital utilization efficiency. For example, when it is predicted that the LCR level will be high in the future, banks can choose to reduce the excessive low-yield liquid assets held and invest in projects with higher returns; on the contrary, when it is expected that the LCR may decline, it is necessary to replenish liquid assets in time to maintain sufficient liquidity buffers. Such flexible adjustments can not only improve the overall profitability of banks but also enhance their ability to cope with market fluctuations.

By incorporating advanced deep learning technology, particularly the GRU network, we aim to significantly enhance the accuracy of LCR predictions, equipping financial institutions with more effective tools for liquidity risk management. This advancement will help individual institutions better navigate potential liquidity crises and support the stability of the entire financial system. Moreover, the research provides financial regulators with new insights and technological approaches to strengthen their supervisory roles and promote a healthy, stable financial market. Applying deep learning to financial risk management also drives technological progress and offers valuable references for addressing similar challenges, fostering innovation in financial technology.

## II. RELATED WORK

Deep learning has increasingly become a powerful tool for solving complex prediction problems in finance, particularly in risk management and time series analysis. Various methodologies have been explored to optimize model performance, which provides a foundation for predicting the Liquidity Coverage Ratio (LCR).

Recent research efforts have demonstrated the efficacy of recurrent neural networks (RNNs) and their variants in processing sequential data, such as time series. The Gated Recurrent Unit (GRU) network, a simplified version of the Long Short-Term Memory (LSTM) network, has shown advantages in capturing long-term dependencies with a more streamlined architecture. This characteristic makes GRUs particularly suitable for applications where computational efficiency and prediction accuracy are critical, such as LCR forecasting. In this context, Liu et al. [9] leveraged Bi-LSTM models with an attention mechanism for optimizing text classification tasks. Although not directly related to financial prediction, their approach to enhancing RNNs with attention mechanisms provides insights into improving GRU-based models by allowing the network to focus on important temporal features.

In the area of financial risk prediction, Wu et al. [10] developed an adaptive feature interaction model to predict credit risk within the digital finance sector. Their study demonstrates the potential of deep learning models in financial applications, indicating that approaches like GRU networks could effectively handle complex, nonlinear patterns in financial data. Similarly, Liu et al. [11] explored the use of Graph Neural Networks (GNNs) in assessing small and medium-sized enterprises' credit risk, showing that advanced deep learning techniques can significantly enhance prediction accuracy in financial contexts. Moreover, the integration of reinforcement learning into neural networks for dynamic fraud detection, as studied by Dong et al. [12], highlights the importance of combining multiple deep learning paradigms to improve model adaptability in rapidly changing environments [13]. This perspective can inspire the development of more robust LCR prediction models that dynamically adjust to new data and evolving market conditions.

Additional works have explored other advanced deep learning techniques, such as contrastive learning [14] and multimodal fusion [15], which can be valuable for enhancing prediction models by incorporating different types of data or optimizing training strategies. These methods, while not directly targeting LCR prediction, provide a broader understanding of techniques that could potentially be adapted to improve GRU models in financial applications.

Furthermore, techniques like spatio-temporal aggregation for fraud risk detection [16] and transforming multidimensional time series into interpretable event sequences [17] emphasize the importance of sophisticated data processing and model interpretability in financial predictions. Applying similar strategies to GRU-based LCR prediction models may enhance their ability to capture important temporal patterns and provide more actionable insights for risk management.

## III. ALGORITHM PRINCIPLE

In order to build a liquidity coverage ratio (LCR) prediction model based on the GRU network, we first defined the basic architecture of the model. The GRU network is a special recurrent neural network that effectively handles long-term dependencies in time series data by introducing update gates and reset gates to control the flow of information. The GRU network architecture is shown in Figure 1. In this study, we will use a multi-layer GRU structure to capture features at different time scales. Assuming that the input time series is $X = \{x_1, x_2, ..., x_T\}$, where each $x_1$ represents a data vector at a time point, our goal is to predict the LCR value at a future time point $T + \Delta T$.

The core of the GRU unit lies in its two gating mechanisms: update gate $z_t$ and reset gate $\tau_t$. These two gating mechanisms are used to determine how to combine new information with historical states and how to forget old information. The calculation formula for update gate $z_t$ is:

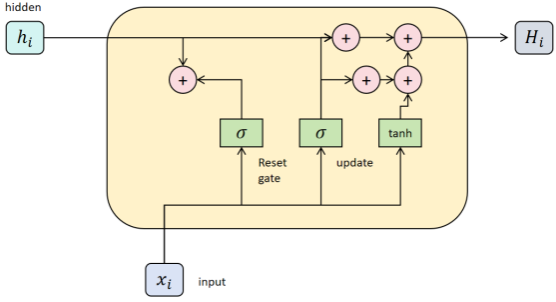

Figure 1 GRU network architecture diagram

$$z_t = \sigma(W_z \cdot [h_t - 1, x_t] + b_z)$$

Where $W_z$ is the weight matrix, $b_z$ is the bias term, and $\sigma$ represents the Sigmoid activation function, which limits the output between 0 and 1, indicating the degree of influence of the current input on the new state. The calculation formula for the reset gate $\tau_t$ is similar:

$$\tau_t = \sigma(W_r \cdot [h_{t-1}, x_t] + b_z)$$

The reset gate determines the importance of the previous state $h_{t-1}$ in calculating the candidate hidden state $h'_t$. The candidate hidden state $z_t$ is calculated by the following formula:

$$h'_t = \tanh(W \cdot [r_t \otimes h_{t-1}, x_t] + b)$$

Where $\otimes$ represents element-by-element multiplication, W and b are weight matrices and bias terms, respectively, and the tanh function is used for nonlinear transformation. The final hidden state $h_t$ is controlled by the update gate $z_t$ and is calculated by the following formula:

$$h_t = (1 - z_t) \otimes h_{t-1} + z_t \otimes h'_t$$

In order to improve the prediction ability of the model, we added a fully connected layer after the GRU network as the output layer, which maps the last hidden state of the GRU to the LCR value. Assuming that the weight matrix of the output layer is $W_o$ and the bias term is $b_o$, the final prediction value $y_T + \Delta T$ can be calculated by the following formula:

$$y_T + \Delta T = W_o \cdot h_T + b_o$$

To train the GRU model, we use Mean Squared Error (MSE) as the loss function, which aims to minimize the difference between the predicted value and the true LCR value. The calculation formula of MSE is as follows:

$$MSE = \frac{1}{N} \sum_{i=1}^{N} (y_i - y'_i)^2$$

During the optimization process, we use the Adam optimization algorithm to adjust the model parameters to achieve the goal of minimizing the loss. The Adam algorithm combines the advantages of momentum gradient descent and adaptive learning rate, and can effectively handle large-scale data sets and accelerate the convergence process. In addition, in order to avoid the overfitting problem, we use the Dropout regularization technique during the training process to randomly discard the output of a part of the neurons to reduce the model's dependence on training data.

## IV. EXPERIMENT

### A. Datasets

In order to conduct the GRU network-based liquidity coverage ratio (LCR) prediction experiment, we chose a widely recognized and publicly available financial dataset, the Federal Reserve Economic Data (FRED). FRED is a huge economic database maintained by the Federal Reserve Bank of St. Louis, which contains a large amount of macroeconomic and financial market data from all over the world. This dataset not only covers a wide range of economic indicators, such as GDP, unemployment rate, inflation rate, etc., but also includes detailed information related to financial institutions, such as bank balance sheets, market interest rates, and various liquidity indicators. The comprehensiveness and reliability of the FRED dataset make it an indispensable resource in financial research.

In this study, we will pay special attention to data related to bank liquidity, including but not limited to reserve requirement ratio, short-term treasury bond yield, money market fund assets, interbank lending rate, etc. These data can reflect the bank's capital flow and market liquidity, and are an important basis for building an LCR prediction model. In addition, FRED also provides rich historical data with a time span of decades, which allows us to use long-term time series data to train and validate our GRU model. By using these high-quality data, we expect to more accurately capture the dynamic changes of liquidity risk and provide financial institutions with effective risk management tools. The openness and diversity of the FRED dataset also provide us with the possibility of multi-angle analysis, which helps to improve the generalization ability and practical application value of the model.

### B. Experimental Results

In order to demonstrate the advancedness of our method, this section selects several mainstream time series forecasting methods for comparative experiments, including autoregressive integrated moving average model (ARIMA), exponential smoothing, long short-term memory network (LSTM), convolutional neural network (CNN) and MLP, and compares these methods with our proposed LCR forecasting model based on GRU network; the evaluation indicators will include mean square error (MSE), root mean square error (RMSE), mean absolute error (MAE) These indicators can comprehensively reflect the prediction accuracy and goodness of fit of the model.

From the experimental results in Figure 2, it can be seen that there are significant differences in the performance of different models in predicting the liquidity coverage ratio (LCR): the MSE value of the ARIMA model is 1.786, which is the highest among all models, indicating that it has limitations in dealing with complex nonlinear and dynamically changing data; the MSE value of the exponential smoothing method is 1.755, which is slightly lower than ARIMA but still high. Although it is suitable for data with trends or seasonality, it performs poorly in financial time series; the MSE value of the multilayer perceptron (MLP) is 1.678, which is an improvement but still insufficient to cope with complex time dependencies; the MSE value of the convolutional neural network (CNN) is 1.696, which is slightly lower than that of the ARIMA model, but still high. The MSE value of is 1.532, which improves the performance through local feature extraction, but has limited ability to process long-range dependencies; the MSE value of the long short-term memory network (LSTM) is 1.473, which performs well and can effectively handle long-term dependency problems, but its structure is complex and the computational cost is high; in contrast, the GRU network-based model we proposed achieved the lowest MSE value of 1.375, which not only performs well in prediction accuracy, but also has a simpler structure and higher training efficiency, which enables GRU to reduce model complexity and computational cost while maintaining high prediction accuracy, thus becoming a very promising method in financial risk management. This result further verifies the superiority of GRU in processing complex financial time series data, and provides strong support for its promotion in practical applications. In order to further demonstrate our experimental results, we provide charts to represent our experimental results.

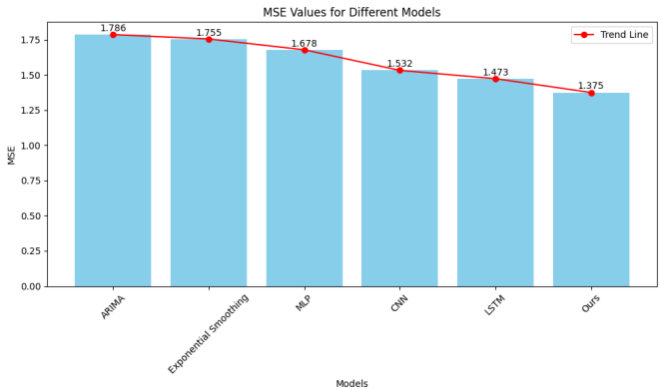

Figure 2 Experimental results of MSE

From the experimental results in Table 1, it can be seen that there are significant differences in the RMSE (root mean square error) performance of different models when predicting the liquidity coverage ratio (LCR). Specifically, the model based on the GRU network proposed by us performs best in the RMSE indicator, with an RMSE value of 1.173, which is significantly lower than several other mainstream time series prediction methods. The RMSE value of the ARIMA model is 1.337, the highest among all models, which shows that it has a large prediction error when dealing with complex financial time series data. The RMSE value of the exponential smoothing method is 1.325, which is slightly lower than ARIMA but still high, showing its limitations in capturing complex nonlinear relationships. The RMSE value of the multilayer perceptron (MLP) is 1.295, which is an improvement but still not ideal, reflecting its shortcomings in dealing with long-term dependencies. The RMSE value of the convolutional neural network (CNN) is 1.238, which further reduces the prediction error, but there is still room for improvement in dealing with long-term dependencies in time series. The RMSE value of the long short-term memory network (LSTM) is 1.214, which is the best-performing model besides the GRU model we proposed, but its structural complexity and computational cost are high. In contrast, the model based on the GRU network not only performs well in the RMSE indicator, but also has a simpler structure and higher training efficiency, making it a very promising financial risk management tool. This result further verifies the superiority of the GRU model in improving the accuracy of LCR prediction. Similarly, we also provide a chart to show our experimental results.

Table 1 Experimental results of RMSE

| Model | RMSE |
| --- | --- |
| ARIMA | 1.337 |
| Exponential Smoothing | 1.325 |
| MLP | 1.295 |
| CNN | 1.238 |
| LSTM | 1.214 |
| Ours | 1.173 |

From the experimental results in Figure 3, it can be seen that the model based on the GRU network (Ours) performs best in terms of mean absolute error (MAE) indicator, with an MAE value of 0.934, which is significantly lower than several other mainstream time series prediction methods. The MAEs of ARIMA and exponential smoothing are 1.065 and 1.054, respectively, showing high prediction errors, which may be due to the limited ability of these two traditional methods to deal with complex nonlinear and dynamic changes. The MAE of the multilayer perceptron (MLP) is 1.032, which is an improvement but still has a lot of room for improvement. The MAE of the convolutional neural network (CNN) is 0.988, which further reduces the prediction error, but its ability to handle long-term dependencies is relatively weak. The MAE of the long short-term memory network (LSTM) is 0.968, which performs well but is slightly higher than GRU. Overall, our GRU model not only performs best in MSE, but also has significant advantages in MAE, indicating that it has higher accuracy and robustness in predicting LCR.

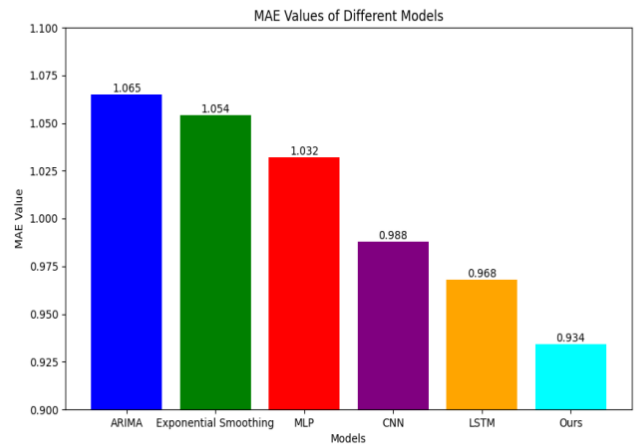

Figure 3 Experimental results of MAE

## V. CONCLUSION

The liquidity coverage ratio prediction model based on the gated recurrent unit (GRU) network has shown great potential in improving the ability of financial institutions to respond to potential liquidity crises. Through the study and analysis of a large number of actual cases, we found that compared with traditional liquidity risk management methods, the use of GRU networks for LCR prediction not only improves the prediction accuracy but also enhances the model's ability to capture the sensitivity of market fluctuations. This means that banks and other financial institutions can optimize resource allocation strategies based on more accurate prediction results. For example, when the future LCR level is expected to be high, they can reduce the holding of too many low-yield liquid assets and invest in projects with higher returns; otherwise, they need to replenish liquid assets in a timely manner to maintain sufficient buffers. Such flexible adjustment measures are conducive to enhancing the ability of individual financial institutions to resist emergencies and contributing to maintaining the sound operation of the entire financial system. In addition, the introduction of advanced deep learning technology, particularly the application of GRU networks, provides new perspectives and technical tools for financial regulators, helping them better perform their supervisory duties and promote the healthy development of financial markets. In summary, by applying the GRU network to LCR prediction, not only can prediction accuracy be significantly improved, but a more effective risk management solution can also be provided for the financial industry, thus helping to build a safer, more stable, and more efficient financial environment.